%% file: main.tex
\documentclass[letterpaper, 10 pt, conference]{ieeeconf}  

\IEEEoverridecommandlockouts                              
                                                          
\usepackage{graphicx} 
\usepackage{amsmath} 
\usepackage{amssymb}  
\usepackage{pseudocode}
\usepackage[ruled]{algorithm2e}
\usepackage{etoolbox}
\usepackage{color,soul}

\makeatletter
\patchcmd{\@algocf@start}
  {-1.5em}
  {0pt}
  {}{}
\makeatother

\usepackage{color}    
\usepackage[export]{adjustbox}
\usepackage[caption=true, font=footnotesize]{subfig}
\usepackage{wrapfig}
\usepackage{tabularx}
\usepackage{booktabs}
\usepackage{multirow}
\usepackage{adjustbox}
\usepackage{pgfplots}
\pgfplotsset{compat=1.9}
\usepackage{lipsum}

\SetAlCapNameFnt{\scriptsize}
\SetAlCapFnt{\scriptsize}

\newcount\cyclecount \cyclecount=-1
\newcount\ind \ind=-1

\newcommand{\etal}{\textit{et~al.}\ }
\newcommand{\eg}{e.\,g.}
\newcommand{\wrt}{w.\,r.\,t.\ }

\definecolor{myblue}{rgb}{0.02,0.27,0.68}

\newcommand{\ii}{(\emph{ii})\ }
\newcommand{\iii}{(\emph{iii})\ }

\long\def\rm#1{\textcolor{red}{#1}}

\usetikzlibrary{external}
\usepgfplotslibrary{groupplots}
\usetikzlibrary{patterns}
\usepackage{filecontents}

\title{\LARGE \bf
Occlusion-Aware Search for Object Retrieval in Clutter
} 

\author{Wissam Bejjani, Wisdom C. Agboh, Mehmet R. Dogar and Matteo Leonetti
		   \thanks{Authors are with the School of Computing, University of Leeds, United Kingdom
        {\tt\small  \{w.bejjani, scwca, m.r.dogar, m.leonetti,\}@leeds.ac.uk}}
}

\begin{document}

\maketitle
\thispagestyle{empty}
\pagestyle{empty}

\captionsetup[figure]{labelfont={bf},name={Fig.},labelsep=period}
\setlength{\textfloatsep}{1.5pt}

\input{0-Abstract}
\input{1-Introduction}

\input{2-RelatedWork}
\input{3-Methodology}
\input{5-Hybrid}
\input{6-Model-free}
\input{7-Experiments}

\input{8.8-Real-World}

\input{9-Conclusion}
\bibliographystyle{IEEEtran} 
\bibliography{main}


\end{document}

%% file: 0-Abstract.tex
\begin{abstract}
We address the manipulation task of retrieving a target object from a cluttered shelf.
When the target object is hidden, the robot must search through the clutter for retrieving it.
Solving this task requires reasoning over the likely locations of the target object.
It also requires physics reasoning over multi-object interactions and future occlusions.
In this work, we present a data-driven hybrid planner
for generating occlusion-aware actions in closed-loop.
The hybrid planner explores likely locations of the occluded target object
as predicted by a learned distribution from the observation stream.
The search is guided by a heuristic trained with reinforcement learning to act on observations with occlusions.
We evaluate our approach in different simulation and
real-world settings (video available on \textcolor{myblue}{https://youtu.be/dY7YQ3LUVQg}). 
The results validate that our approach can search and retrieve a target object in near real time in the real world while only being trained in simulation.
\end{abstract}

%% file: 1-Introduction.tex
\section{Introduction} \label{sec:INTRO}
\begin{figure*}
\centering
\includegraphics[max width=1.0\textwidth]{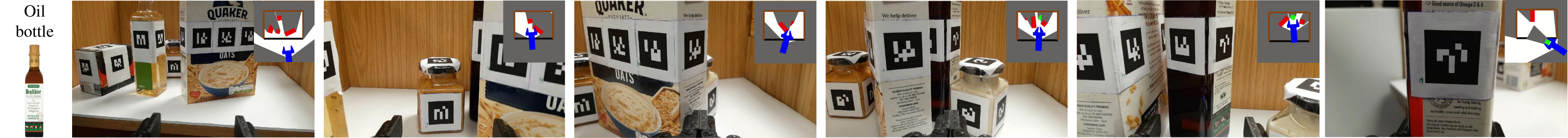}
\caption{The robot is tasked with retrieving the oil bottle. The real-world images are from a gripper-mounted RGB camera. The images on the top-right are rendered from the current state of the physics simulator as recreated based on the current real-world image. The images from the simulator are used by our approach to hypothesize potential poses for the target object.
}
\label{fig:seq1}
\end{figure*}

Autonomously
manipulating everyday objects 
in cluttered environments with occlusions has
long been a target milestone in robotics research \cite{roadmap2020, christensen2016roadmap}.
As an example scenario consider Fig.~\ref{fig:seq1}, in which  the robot is tasked with retrieving 
the oil bottle from a kitchen cabinet of limited height.
The cabinet shelf is cluttered with with jars, cereal boxes, and other bottles
while the oil bottle is nowhere to be seen. 
The robot needs to push through the clutter 
to search for the oil bottle, and then reach, grasp, and pull it out
without dropping any of the other objects off the 
shelf.

A sequence of prehensile and non-prehensile actions 
in a partially observable and contact-rich environment requires 
reasoning on occlusions and physics-based uncertainty. 
Even when high-accuracy object detection systems are available,
occlusion remains an inherent source of uncertainty hindering the 
search for the target object \cite{kaelbling2013integrated}.
The robot has to reason over a history of partial observations to efficiently
explore where the target object might be.
Furthermore, it is notoriously hard to predict the outcome of 
an action in multi-contact physics environments 
\cite{ronnau2013evaluation, chung2016predictable, leidner2018cognition, papallas2020non}.
Modelling error on the physics parameters such as friction, inertia, and objects shapes 
impede open-loop execution of long action sequences.

Most research efforts on sequential-decision making
in clutter and under partial observability 
have focused on model-based approaches. 
When the task is modelled as a
Partially Observable Markov Decision Process (POMDP)~\cite{kaelbling1998planning},
planning takes place in belief space,
that is, on a probability distribution over the actual state.
The belief is continuously updated after every interaction with the environment~\cite{li2016act, xiao2019online, pajarinen2017robotic}.
In multi-contact multi-object tasks, however, 
the physics can quickly degenerate to multi-modal 
and non-smooth distributions~\cite{agboh2020parareal}.
Hence, scaling the belief update over occluded spaces and the belief planner to
long action sequences become impractical.
Alternatively, model-free approaches with function approximators
bypass the need for a closed-form representation 
of the belief update and environment dynamics.
By directly mapping observation history to manipulation actions,
they can scale to arbitrary large state spaces and with long observation history
\cite{garg2019learning, heess2015memory, danielczuk2019mechanical}.
Sequential reasoning over future occlusions and multi-contact physics remains an open challenge
for model-free approaches.

To solve the problem of multi-object manipulation under physics uncertainty, 
heuristic-guided Receding Horizon Planning, RHP, can be used. 
RHP interleaves quick short horizon planning cycles with execution, similar 
to model predictive control.
By continuously updating the simulator state, where planning takes place, 
from real-world observations, RHP circumvents the problem of compounding modelling errors over long sequences of actions.
Under the assumption of a fully observable environment, 
we have shown in our previous work how RHP can be used with
a heuristic
to guide physics-based roll-outs and to estimate the cost-to-go from 
the horizon to the goal \cite{bejjani2018planning}.
This approach balances the advantages of model-based sequential reasoning 
with a model-free scalable heuristic 
\cite{bejjani2019learning, bejjani2021learning}.
However, in a partially observable environment,
the target object is not always detected and hence cannot be simulated by RHP.
In this work, we explore learning to predict the location of 
the target object.

We propose 
(\emph{i}) a data-driven approach for maintaining a distribution
over the target object's pose from a stream of partial observations
\ii and an occlusion-aware heuristic 
to run RHP under partial observability.
These two key ideas form a hybrid planner which uses the distribution to suggest 
potential target object poses
for RHP to explore.
We also present the learning architecture for simultaneously learning a generative model 
of the target object pose distribution and an occlusion-aware heuristic in a continuous action space.
We evaluate the proposed approach in environments 
with varying clutter densities, configurations, and object shapes. 
We also validate its performance in retrieving 
different target objects in the real world.
A holistic analysis of these contributions can be found in \cite{bejjani2021phd}.

This work adopts the following assumptions.
A library of object type-shape pairs is given.
Objects have a uniform horizontal cross-section along the z-axis, 
and they are small enough 
to be graspable from at least one approach angle. 
They are placed on the same horizontal 
surface.
The robot's actions are pose increments
parallel to the manipulation surface 
in the planar Cartesian space of the gripper.
We do not consider 
access to a separate storage space.

%% file: 2-RelatedWork.tex
\section{Related Work} \label{sec:SOTA}
\emph{\textbf{POMDP planners:}} In the presence of occlusions, manipulation in clutter 
is often associated with active search, that is,
leveraging manipulation actions to simultaneously 
gain visibility and accessibility \cite{bohg2017interactive}.
Thanks to recent advances in model-based online planners under uncertainty \cite{somani2013despot, xiao2019online, papallas2020online, bejjani2015automated}, 
this field is gaining momentum towards achieving everyday 
manipulation tasks.
Wong \etal \cite{wong2013manipulation} 
use object semantics and spatial constraints to focus the search in shelves where the clutter is
most similar to the target object.
Pajarinen \etal \cite{pajarinen2017robotic} solve long-horizon multi-object 
manipulation by combining particle filtering  
and value estimates in an online POMDP solver.
These approaches 
have largely overcome the computational 
complexity 
associated with 
large state space and observation history. 
However, they avoid multi-object contacts by 
planning with collision-free actions. 
This constraint reduces planning time, but it also prevents 
the robot from exploiting the full dynamics of the domain.


\emph{\textbf{Model-free policies with recurrent units:}} 
Model-free policies are at the core of many applications that 
necessitate reactive decision-making under uncertainty.
Heess \etal \cite{heess2015memory} show that by using 
Long Short-Term Memory (LSTM) cells as a tool to summarize 
a history of partial observations, 
it is possible to train a policy for pushing an object to an initially observed pose. 
Karkus \etal \cite{karkus2017qmdp} propose a model-free approach that trains a neural network (NN)
on expert demonstrations to approximate a Bayesian filter and a POMDP planner.
These approaches are focused on single object manipulation and do not ensure
long-term reasoning over the physics.

\emph{\textbf{Searching in clutter through manipulation:}}
The goal of our work is most aligned with the objective of
Danielczuk~\etal~\cite{danielczuk2019mechanical}. 
They define it as
``Mechanical Search'', a long sequence of actions for 
retrieving a target object from a cluttered environment 
within a fixed task horizon while minimizing time.
They propose a data-driven framework for detecting then performing 
either push, suction, or grasp actions until the target 
object is found.
They tackle top-down bin decluttering by removing obstructing objects
until the target is reachable.
Such an approach requires a separate storage space to hold obstructing objects. 
To address environments where a separate storage space is not available,  
Gupta~\etal~\cite{gupta2013interactive} 
and Dogar~\etal~\cite{dogar2014object}
interleaves planning
with object manipulation on a shelf.
They both propose moving objects to unoccupied spaces within the same shelf
to increase 
scene visibility from a fixed camera view angle.
The approaches stated so far perform the search by
manipulating one object at a time, 
avoiding sequential reasoning over multi-contact physics.
Avoiding 
all obstacles 
remains, however, impossible (and often undesirable) in many 
partially observable and cluttered environments.
Most recently, Novkovic \etal \cite{novkovic2020object} propose a 
closed-loop decision making scheme for 
generating push action in a 
multi-contact physics environment
with a top-mounted camera. 
Their approach relies on encoding the observation history in 
a discretized representation of the environment. 
The encoding is used by an RL trained policy to generate 
the next push action for revealing hidden spaces. 
We adopt a similar decision making scheme,
but we avoid the limitations
of encoding the observation history in a discretized representation.
Instead, we rely on the NN's recurrent
units to capture the observation history.

%% file: 3-Methodology.tex
\section{Problem Definition} \label{sec:METHO}
The robot's task is to retrieve a target object from a shelf 
of limited height
without dropping 
any of the other objects off the shelf.
The robot carries a gripper-mounted camera.
We treat the search, reach, grasp, and pull-out of the target object as 
a single optimization problem with the objective of minimizing the total number of actions
for retrieving the target object.

\subsection{Formalism} \label{sec:FORM}
We model the problem as a POMDP  $\langle S, A, O, T, \Omega, r, \gamma \rangle$, where
$S$ is the set of states, $A$ the set of continuous actions, $O$ the set of possible observations, $T : S \times A \times S     \rightarrow [0,1]$  the transition function,  $\Omega : S \times O \rightarrow [0,1]$ the observation model, 
$r:  S \times A \times S \rightarrow  \mathbb{R}$ is the reward function, and $\gamma \in [0,1)$ is the discount factor. 
$s=\{Rob, Obj_1, Obj_2, \ldots  \}$,
in which 
$Rob$ is the robot's end-effector pose, shape, and gripper's state; 
$Obj_i$ describes an object's pose, shape, and type.
An observation $o \in O$ contains a subset of the state variables (\eg, the visible objects), and the geometry of occluded spaces: the shadowed areas behind objects and areas outside the
camera's field of view (FOV). 

Since the state is not always accessible because of occlusions, 
decision making relies  on maintaining a belief $b : S \rightarrow [0,1]$ as a distribution over possible states.
A POMDP policy $\pi$ is a function that maps a belief $b$ to an action $a$. 
The value $V$ of a policy $\pi$ 
at belief $b_t$ at time $t$ is the expected \emph{return}:
$
V_{\pi} = \mathbb{E}_{a\sim\pi, s_t \sim b_t} [\sum_{k=t} \gamma^{k-t} r_{k+1}] 
$
where $r_{t+1} = r(s_t,a_t, s_{t+1})$.
We avoid shaping the reward function in order not to skew the robot's behaviour towards 
any preconceived human intuition which might artificially limit the return. 
Instead, we opt for a constant 
negative reward of $-1$ per action. 
When an object is dropped, the task is terminated and 
an additional large negative reward of $-50$ is received.

\subsection{Overview} \label{sec:OVER}
\begin{figure}[!t]
  \centering
  \includegraphics[width=0.5\textwidth]{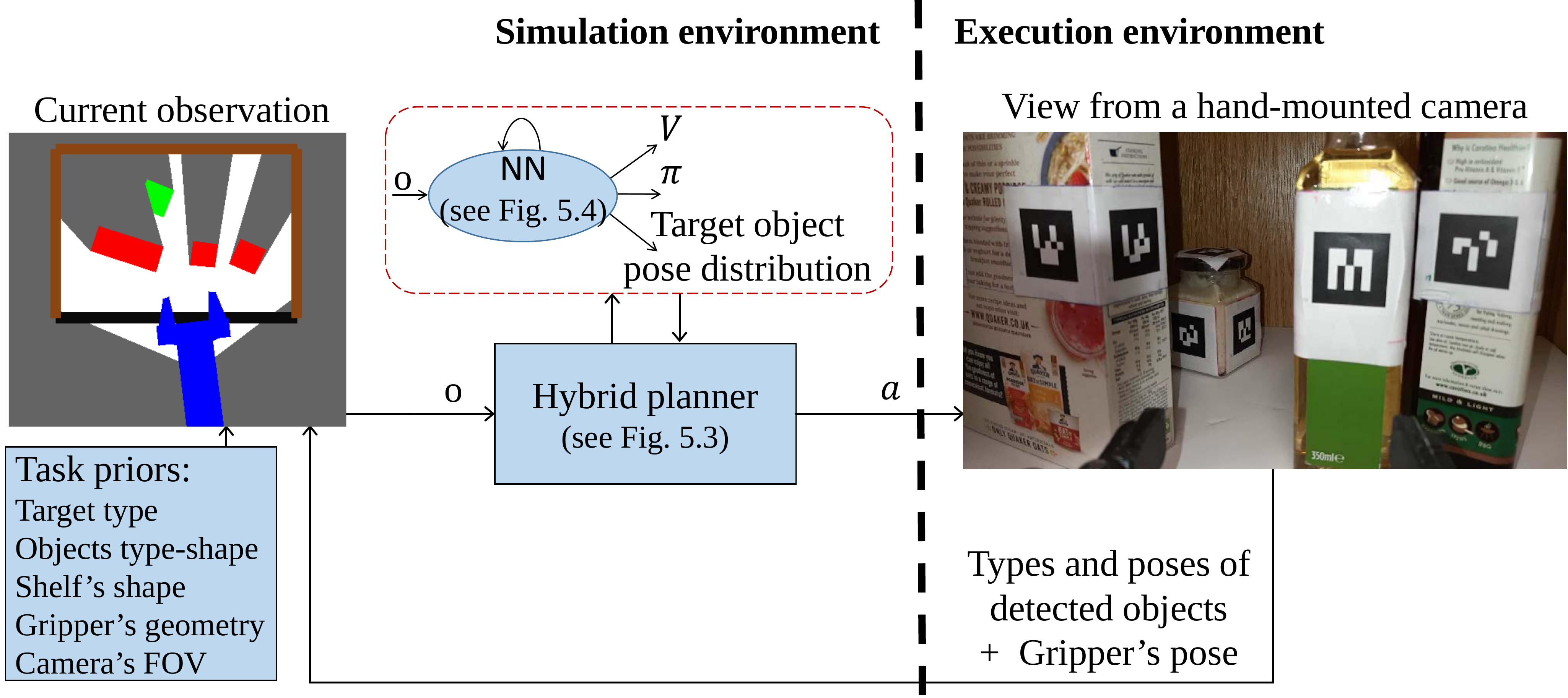}
  \caption[]{Approach overview. An example with the small jar at the back of the shelf as the target object.}
  \label{fig:control}
\end{figure}

We use the closed-loop decision making scheme shown in Fig.\ref{fig:control},
where we observe the environment, plan, execute the first action of the plan, then loop back to the observe step.\\
\emph{\textbf{Observe}}: 
The poses and types of visible objects in the execution environment,
as detected by the gripper-mounted camera,
and task priors are used to 
recreate, in the simulation environment, a state with the currently detected objects.
The current observation,
a top-down view of the scene,
is rendered from the simulation
environment (Sec.\ref{sec:OBSER}). 
\\
\emph{\textbf{Plan}}:
The hybrid planner uses 
the observation history, including the current observation,
to update a distribution over the likely poses of the target object.
The estimated target object poses are used to hypothesize 
root states, each with a target object placed at one of the predicted poses.
If a predicted pose is in an occluded area, 
the target object would still be added to the physics simulator but it will be occluded in the observation.
RHP uses its occlusion-aware heuristic (a stochastic policy and its value function) 
to explore and evaluate physics roll-outs from the root states.
RHP returns the best action to execute at each root state 
and its corresponding estimated \emph{return} (Sec.\ref{sec:HYB}).\\
\emph{\textbf{Execute}}:
The \emph{returns} are weighted by the likelihood of their root states, 
and the action with the highest weighted \emph{return} 
is executed in the execution environment (Sec.\ref{sec:HYB}).
After a single step of execution, the system goes back to the observation step,
for a closed-loop execution.
    

At the core of our approach is a NN with recurrent units that maps 
an observation history into: 
(\emph{i}) a distribution over the pose of the target object $\hat{y}(\Bar{o})$ with $\Bar{o}$ being the observation history,
\ii a stochastic policy $\pi(.|\Bar{o})$, 
\iii and its corresponding value function $V_{\pi}(\Bar{o})$, 
(Sec.~\ref{sec:MOD_FREE}). 
The NN is trained in the 
physics simulation environment 
with 
curriculum-based Reinforcement Learning (RL) (Sec.~\ref{sec:MOD_FREE}).

%% file: 5-Hybrid.tex
\section{Decision Making Under Occlusion} \label{sec:OPC}
\subsection{Observation Space} \label{sec:OBSER}
It is essential to have an expressive representation of the observation
yet compact enough to keep the NN size relatively small as it will 
be queried multiple times per action selection.
Even though in the real world the camera is gripper-mounted, 
before we feed the observation into the NN,
we render it in a top-down view, 
as shown in the top-left of Fig.\ref{fig:control},
making
the spatial relationships between objects and the geometry of 
occluded and observable areas more explicit.

We built on the abstract image-based representation 
of a fully observable environment in \cite{bejjani2019learning, bejjani2021learning}.
In addition to colour labelling objects based on their functionality (target in green, clutter in red, and surface edges in black),
we represent occluded and observable spaces by white and grey coloured areas respectively. 
The geometry of the occluded areas is computed by illuminating 
the scene from the robot's camera perspective.
We use a black line to represent the shelf edge and brown 
for the shelf walls. 
The top-down view enables data from the execution environment and task priors to be combined.
\begin{itemize}
    \item Object detection in the execution environment 
    identifies the poses and types of visible objects in the camera FOV.
    The objects' poses and types allow the simulation environment to place the correct object shape and colour in the abstract image-based representation of the observation.
    The pose of the robot's gripper is computed from the robot forward kinematics.
    \item The task priors consist of observation-invariant information:
    the type of the target object, the shape corresponding to every object type,
    the shape of the shelf (walls and edge), 
    the geometry of the gripper, 
    and the camera FOV.
    By including task priors in the representation, 
    the learner does not need to remember them from the observation stream.
\end{itemize}


\subsection{Hybrid Planner} \label{sec:HYB}
\begin{figure}[t]
  \includegraphics[width=0.5\textwidth]{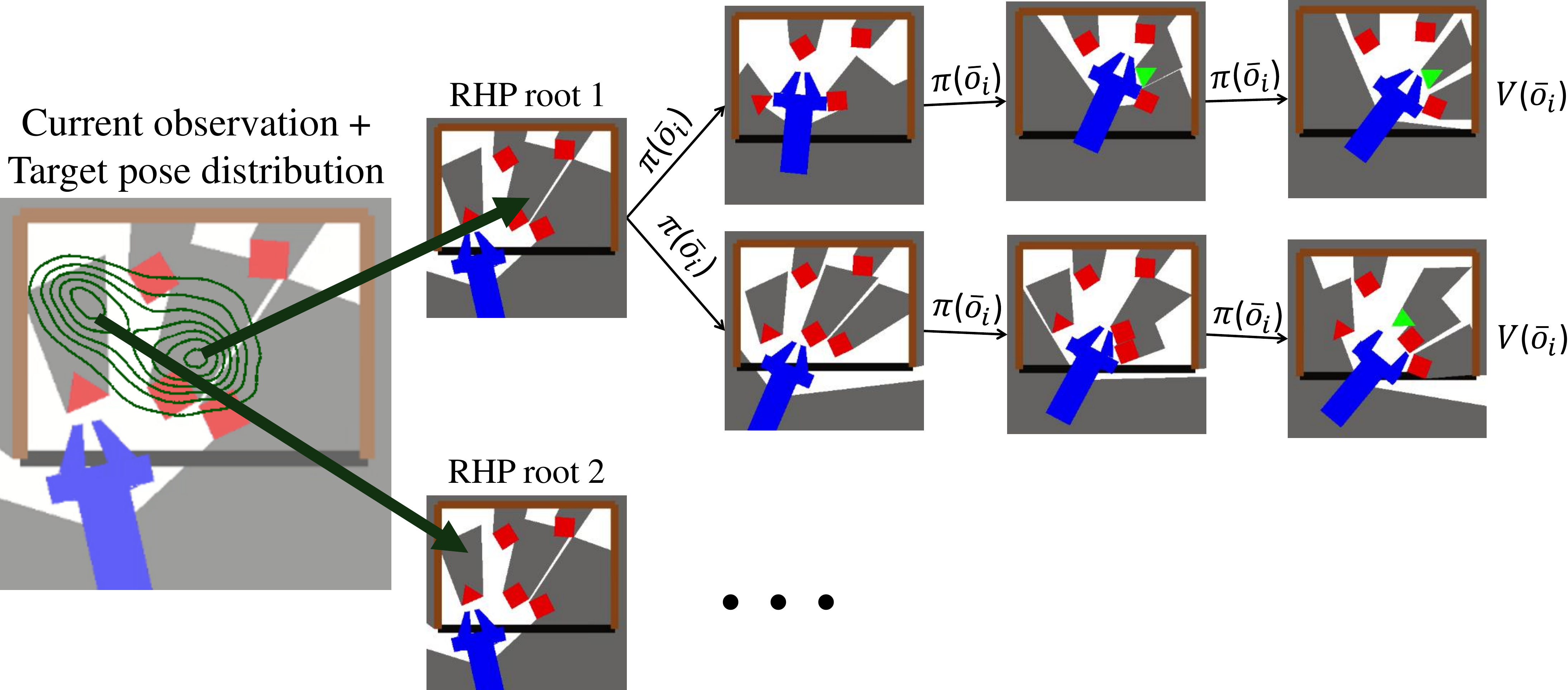}
  \caption{Hybrid planner running two RHP queries,
  one for each peak represented by the contour lines (left).
  RHP is shown executing $2$ roll-outs of depth $3$ for each root state.}
  \label{fig:HYB}
\end{figure}
The hybrid planner algorithm, presented in Alg.\ref{alg:HYB} and 
illustrated in Fig.~\ref{fig:HYB}, is detailed as follows:\\
\textbf{State Generation} (Alg.~\ref{alg:HYB}, line~\ref{l:genState}):
With information from previous observations
captured in the NN recurrent units,
the NN uses the current observation 
to generate an updated distribution over
target object pose. 
For each peak in the distribution,
the hybrid planner creates a root state with the target object placed at the peak location,
while the obstacle poses remain the same as in the current observation. 
The weight of a root state is
computed as the relative likelihood 
of its corresponding peak. 
It measures how likely it is for the target object to be found 
at the predicted location compared to the other potential sites.
RHP is then called over each of the root states (Alg.~\ref{alg:HYB}, line~\ref{l:queryRHP})
\\
\textbf{Occlusion-aware RHP} (Alg.\ref{alg:RHP}):
RHP performs $m$ stochastic roll-outs from root state $s_0$
up to a fixed horizon depth $h$ in the physics simulator. 
Each roll-out is executed by following the stochastic policy $\pi(\Bar{o})$ acting on the observation
history.
The \emph{return} $R_{0:h}$ of a roll-out is computed as the sum of the discounted rewards 
generated by the model and the expected return beyond the horizon estimated by the value function $V(\Bar{o}_h)$:
$
R_{0:h} =  r_1 +  \gamma r_2 + \ldots + \gamma^{h-1}r_{h} + \gamma^{h}V(\Bar{o}_h).
$
RHP returns the \emph{first} action $a_0$ and $R_{0:h}$ of the roll-out 
that obtained the highest \emph{return}. \\
\textbf{Action Selection} (Alg.~\ref{alg:HYB}, line~\ref{l:actSel}):
The \emph{return} of every RHP query is scaled by the weight of its root state (Alg.~\ref{alg:HYB}, line~\ref{l:scale}). Therefore, the robot picks the action that maximizes the \emph{return} with respect to both the probability of the roll-out, and the probability of the location of the target object.
\begin{algorithm}[t]
  \scriptsize
  \LinesNumbered
  \SetKwInput{Input}{Input}
  \SetKwInOut{Output}{Output}
  \Input{trained neural network NN,
        observation history $\Bar{o}$, \qquad\qquad\qquad\qquad
        number of roll-outs  $m$, horizon depth $h$}
  \Output{action $a_r$}
    \textit{rootActions} $\leftarrow$ [ ], \, \textit{weightedReturns} $\leftarrow$ [ ] \\
    \textit{rootStates}, \textit{rootWeights} $\leftarrow$ generateStates(NN, $\Bar{o}$)\label{l:genState}\\
    \For{$s_o, w$ in [rootStates, rootWeights]}{
        $a_{r}, R_{0:h}$ $\leftarrow$ RHP(NN, $s_o$, $\Bar{o}$, $m$, $h$)\label{l:queryRHP}\\
        \textit{rootActions}.append($a_{r}$) \\
        \textit{weightedReturns}.append($w \times R_{0:h}$) \label{l:scale}\\
    }
    \textbf{return} rootActions[$argmax$(\textit{weightedReturns})] \label{l:actSel}
    \caption{Hybrid planner (NN, $\Bar{o}$, $m$, $h$)}
    \label{alg:HYB}
\end{algorithm}
\begin{algorithm}[t]
  \scriptsize
  \SetKwInput{Input}{Input}
  \SetKwInOut{Output}{Output}
  \Input{
         trained neural network NN,
         root state $s_0$, 
         observation history $\Bar{o}$, \qquad \qquad
         number of roll-outs  $m$, horizon depth $h$}
  \Output{action $a_0$, return $R$}
    $RolloutsReturn \leftarrow$ [ ], \,
    $FirstAction \leftarrow$ [ ] \\
    \For{$i$ = 1,2, \ldots, $m$}{
        $R \leftarrow 0$, \, $\Bar{o}_i \leftarrow \Bar{o}$ \\
        $s, \ o \leftarrow $ setSimulatorTo$(s_{0})$ \\
        $\Bar{o}_i$.append$(o)$\\
        \For{$j$ = 1,2, \ldots, h}{
            $a \sim \pi(.|\Bar{o}_i)$  \\      	
            \If{$j$ \textbf{is} 1}{
                $FirstAction$.append($a$)
            }
            $s, \ o, \ r \leftarrow$ simulatePhysics($s,a$) \\
            $R \leftarrow R + \gamma^{j-1}r$ \\
            $\Bar{o}_i$.append$(o)$\\
            \lIf{isTerminal($s$)}{ 
                \textbf{break}
            }
          }
        \lIf{\textbf{not} isTerminal($s$)}{ 
            $R \leftarrow R + \gamma^{h}V(\Bar{o}_i)$
        }
        $RolloutsReturn$.append($R$)
    }
    \textbf{return} $FirstAction[$argmax$(RolloutsReturn)]$,  max$(RolloutsReturn)$
    \caption{RHP (NN, $s_o$, $\Bar{o}$, $m$, $h$) with an occlusion-aware heuristic}
    \label{alg:RHP}
\end{algorithm}

%% file: 6-Model-free.tex
\section{Training the Three-Headed NN} \label{sec:MOD_FREE}
\begin{figure}[]
  \includegraphics[width=0.48\textwidth]{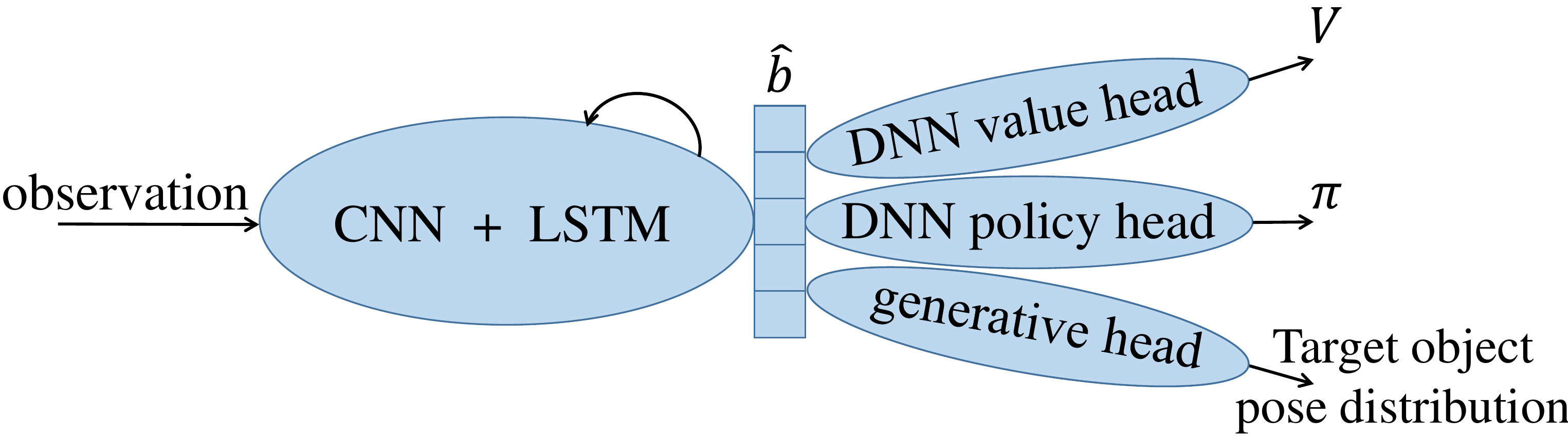}
  \caption{NN architecture.}
  \label{fig:NN}
\end{figure}

Prior to using the NN in the closed-loop decision making scheme,
the NN is trained in a physics simulation environment
(the same environment that will be used by the hybrid planner).
The NN must
(\emph{i}) generalize over variable number of objects and shapes in the observations,
\ii and maintain a belief from the observation stream
in order to 
predict the distribution over the target object pose 
and to generate an informed search and retrieve policy and value function
for RHP to use them as a heuristic.
The NN architecture that satisfies these conditions is illustrated in Fig.\ref{fig:NN}.
The first two components are a Convolutional Neural Network (CNN)
connected to LSTM units.
The CNN takes advantage of having an abstract image-based representation of the observation
to ensure generalization over object shapes and numbers.
The output of the LSTM layer, $\hat{b}$, summarizes the stream of CNN embeddings
into a latent belief vector. 
$\hat{b}$ is then passed through a feed-forward Deep Neural Network (DNN) 
that models the policy, another DNN for the value function, and 
a generative head 
for estimating the pose distribution
of the target object.
The generative head outputs a heat-map, $\hat{y}$, of size equal to the input image, 
where higher pixel values indicate higher chances that the target object is at that location.
As it is common to have the policy and value function sharing
some of NN parameters to stabilize the learning
\cite{mnih2016asynchronous, schulman2017proximal}, 
we also found that having the generative head sharing the CNN and LSTM components 
of the NN with the policy and value function acts as a regularizing element. 

Training a randomly seeded $\theta$-parametrized
NN with recurrent units over images
in a partially observable environment with complex physics 
and in a continuous actions space 
is particularly challenging \cite{mohammed2020review}. 
To increase the likelihood of convergence,
the learning algorithm
uses
RL with a curriculum \cite{narvekar2020curriculum}.
The curriculum is constructed over three task parameterizations to gradually increase the clutter density and, by consequence, 
the occlusion in the environment.
The first parameterization
consists of environments with
random number of objects between $1$ and $4$.
The initial poses of the
target and clutter objects are sampled from a uniform distribution over the shelf. 
The next task parameterization uses between $5$ and $10$ objects.
The final task parameterization limits the minimum number of objects to $7$ 
and the pose of the
target object is sampled from a uniform distribution covering only the back half of the shelf.
Throughout the training, we use random polygon-shaped objects 
for the NN to learn generalizable features. 

The policy and the value function are trained 
with synchronous Advantage Actor-Critic (A2C) \cite{wu2017scalable}.
The generative head is trained in a supervised fashion.
The target $y$ for updating the generative head is  
a heat-map
showing the ground 
truth pose of the target object as given by the simulator. 
The combined loss function is, therefore:
\begin{align*}
& \mathcal{L}_{\textrm{}}(\theta) =
  \frac{1}{M} \sum_{i=1}^{M} 
    -  Adv(\Bar{o}_i,r_i,\Bar{o}'_i)  \;  \textrm{log}\pi_{\theta}(a_i|\Bar{o}_i)  \nonumber \\
    & +  c_1 \;  (r_{i}+\gamma V_{\theta_{old}}(\Bar{o}'_i)-V_{\theta}(\Bar{o}_i))^2 \nonumber \\
    & -  c_2 \;  H(\pi_{\theta}(.|\Bar{o}_i)) \nonumber \\
    & -  c_3 \; \frac{1}{jk}\sum_{j,k}( y^{jk}_i\textrm{log}\hat{y}_\theta^{jk}(\Bar{o}_i) +  (1-y^{jk}_i)\textrm{log}(1-\hat{y}_\theta^{jk}(\Bar{o}_i)),
\label{eq:update_both}
\end{align*}
where $c_1$, $c_2$, and $c_3$ are hyper-parameters, $M$ is the batch size,
$H$ is the entropy term added to encourage exploration,
$j$ and $k$ are the heat-map pixel indices,
and 
$Adv$ is the advantage function estimate formulated over the observation history:
\begin{equation*}
Adv(\Bar{o}_i, r_i, \Bar{o}'_i) = r_{i} +\gamma V_{\theta_{old}}(\Bar{o}'_i)-V_{\theta_{old}}(\Bar{o}_i).
\end{equation*}



%% file: 7-Experiments.tex
\section{Experiments} \label{sec:EXP}
We ran a number of experiments in a physics simulator
and in the real world.
The goals of the experiments are two-fold: 
(\emph{i}) to evaluate the performance of the proposed approach in dealing with occlusion and physics uncertainties,
\ii to verify the approach's transferability to retrieve different target objects in the real world.

\subsection{Evaluation Metrics}\label{sec:EV_MET}
We select evaluation metrics that allow us to 
quantitatively measure the aforementioned goals.
(\emph{i}) The first metric is \emph{success rate}. 
A task is considered successful if the target object is retrieved 
in under $50$ actions, the total task planning and execution time 
is under $2$ minutes, and none of the objects
are dropped off the shelf.
\ii As we also target real-time applications, 
the second metric is the 
\emph{average planning and execution time per task}.
\iii The \emph{average number of actions per task} is the third 
metric as the learning objective 
is to solve the problem with the minimum number of actions.

\subsection{The hybrid Planner and Baseline Methods}\label{sec:}
\textbf{Hybrid planner}: The NN is trained as in Sec.~\ref{sec:MOD_FREE}.
It takes a $64$$\times$$64$$\times$$3$ input 
image\footnote{We used robot-centric images, 
i. e., the colour-labelled abstract images track the
robot from the top-view perspective. 
We found that the robot-centric view reduces
the amount data required by the learning algorithm due to the symmetry of the scene when compared to a world-centric view.}. 
The CNN is composed of three consecutive layers of 
convolution, batch normalization, and maxpooling. 
We use $8$, $8$, $16$ filters of size $3$$\times$$3$ and strides $2$$\times$$2$. 
The CNN is followed by a single LSTM layer of $128$ units.
The policy head is composed of two dense layers with $128$ neurons each. 
The policy output layer has $8$ neurons corresponding to the means 
and standard deviations of the horizontal, lateral, rotational, and gripper actions. 
We use $tanh$ activation function for the means and $sigmoid$ for the standard deviation.
The value head has two dense layers with $128$ and $64$ neurons respectively, 
and a single neuron for the output with linear activation function.
The generative head follows a sequence of three upsampling and convolution layers. 
The filter sizes are 8, 8, 16 and $3$$\times$$3$. 
The final layer is a $64$$\times$$64$$\times$$1$ convolution layer with 
linear activation function followed by a 
$sigmoid$ function to decode the heat-map.
Except for the output layers, we use a $leaky\ relu$ activation throughout the network.  
The NN is updated using the RMSProp optimizer in TensorFlow \cite{tensorflow2015-whitepaper}. 
We use the PPO formulation for the policy loss function \cite{schulman2017proximal}. 
Following the proposed training curriculum, the difficultly of the task is increased every time the success rate of the hybrid planner with $m$=$4$ and $h$=$4$ exceeds $80\%$. 
The training is terminated once the success rate converges. 
We use the following learning parameters: 
$learning\ rate$=$0.00005$, $c_1$=$0.5$,
$c_2$=$0.01$, $c_3$=$1.0$, $\gamma$=$0.995$, 
and $M$=$1500$.
We compare three versions of the hybrid planner with $m$ and $h$ RHP parameters of $2$$\times$$2$, $4$$\times$$4$, and $6$$\times$$6$.
\\
\textbf{Hybrid planner limited}: Instead of performing weighted evaluations of multiple RHP queries, 
this baseline only evaluates the most likely target pose and executes the predicted action.
We implement it with $m$=$4$ and $h$=$4$.
\\
\textbf{Greedy}: This policy presents a deterministic model-free approach. 
The NN is trained similarly to our approach 
excluding the generative head from the architecture.
The robot is directly controlled by the policy head of the NN (without RHP). 
Actions are defined by the mean of the action distribution outputted by the 
policy head over the continuous planar actions space.
It is inspired by 
\cite{novkovic2020object}.
\\
\textbf{Stochastic}: 
This policy is a stochastic version of the greedy policy. 
Actions are sampled from the policy output.
As shown in \cite{jaakkola1995reinforcement},
RL trained stochastic policies provide higher \emph{return} than deterministic ones in a POMDP.
\\
\textbf{Stochastic$_{\mathrm{gen}}$}: We also evaluate an 
additional stochastic policy that samples the policy head
of the NN trained with the generative head.
The purpose is to investigate if 
the policy learns a better reasoning
about the target object pose distribution
when trained using our proposed approach.
\\
\textbf{Hierarchical planner}: This approach offers a model-base baseline.
The low level plans are generated either with kinodynamic RRT \cite{haustein2015kinodynamic} or following a hand-crafted heuristic.
The low level plans are executed in open-loop.
The high level planner has access to the following actions: 
Search( ): positioned outside the shelf,
the robot moves from the far left to the far right of the shelf while pointing the camera inwards.
Throughout this motion, information is collected
on the pose and type of detected objects.
Rearrange($Obj_i$): move a certain object to a free-space in the back of the shelf
by planning 
with Kinodynamic RRT on collected information from the previous Search action.
Move$\_$out( ): rotates the robot to face the inside of the shelf, 
then moves the robot out following a straight line heuristic. 
Retrieve($Obj_i$): plan with Kinodynamic RRT on available information to reach, grasp, and pull-out the target object. 
The high level planner is outlined in Alg.~\ref{alg:HIE}.
This baseline is an adaptation of \cite{dogar2014object}.
\begin{figure}
\captionsetup[subfloat]{farskip=-2pt,captionskip=-4pt}
\captionsetup[subfigure]{labelformat=empty}
\hspace*{-1.1em}
\subfloat[]{ \includegraphics[width=0.162\textwidth]{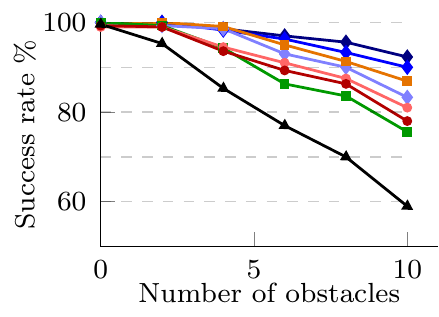} }\hspace*{-0.45em}
\subfloat[]{ \includegraphics[width=0.162\textwidth]{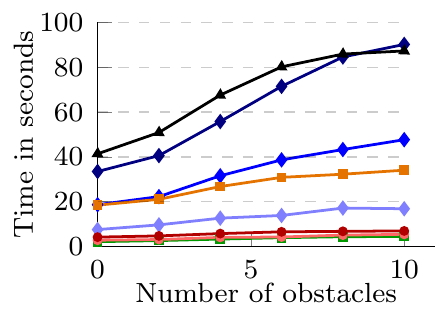}  }\hspace*{-0.45em}
\subfloat[]{ \includegraphics[width=0.162\textwidth]{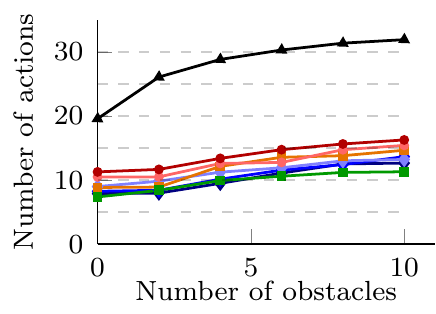} }

\hspace*{-1.1em}
\subfloat[]{ \includegraphics[width=0.162\textwidth]{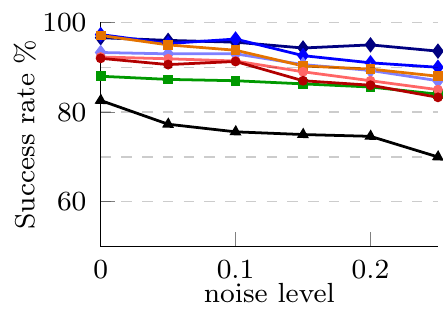}  }\hspace*{-0.45em}
\subfloat[]{ \includegraphics[width=0.162\textwidth]{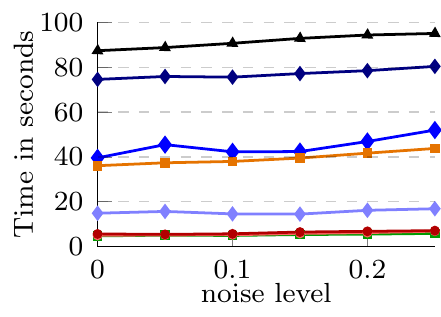}   }\hspace*{-0.45em}
\subfloat[]{ \includegraphics[width=0.162\textwidth]{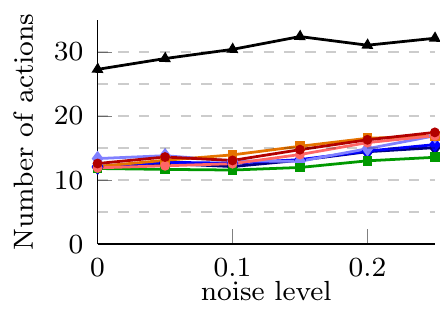}  } 

\hspace*{-0.7em}
\subfloat[]{ \includegraphics[width=0.495\textwidth]{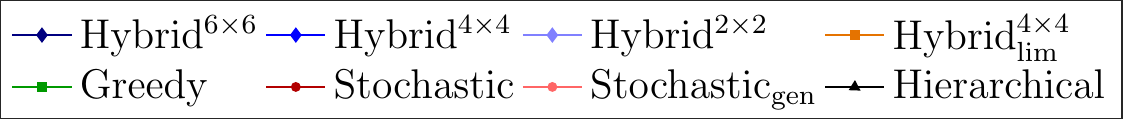}  }

\caption{Performance \wrt different clutter densities and noise levels.}
\label{exp:sim}
\end{figure}

\begin{algorithm}[t]
\scriptsize
\While{target object \textbf{not} retrieved}{
    Search( )\\
    \eIf{target object \textbf{not} located}{
        Rearrange(\textit{closest object to robot})\\
        Move$\_$out( )\\
        }
    {
        Retrieve(target object)
        }
}
\caption{Hierarchical planner}
\label{alg:HIE}
\end{algorithm}

\subsection{Simulation Experiments}
\begin{figure}[b]
  \centering
  \includegraphics[width=0.49\textwidth]{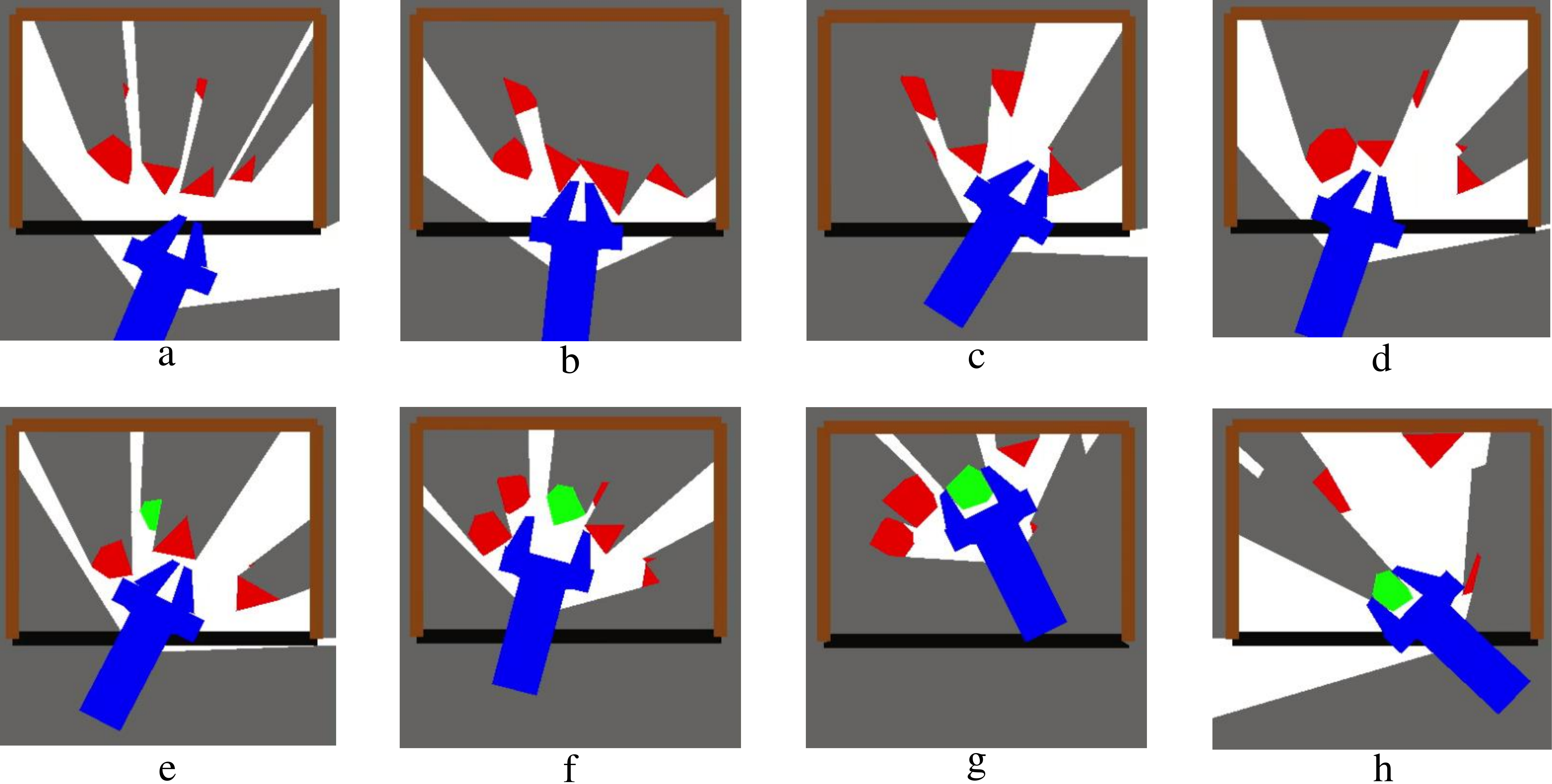}
  \caption{Snippets of the current observation with noise level=0.15. Task solved with Hybrid$^{4\times4}$.}
  \label{fig:noiseSeq}
\end{figure}

\begin{figure*}[!t]
\captionsetup[subfigure]{labelformat=empty}
\subfloat[]{ \includegraphics[width=1.0\textwidth]{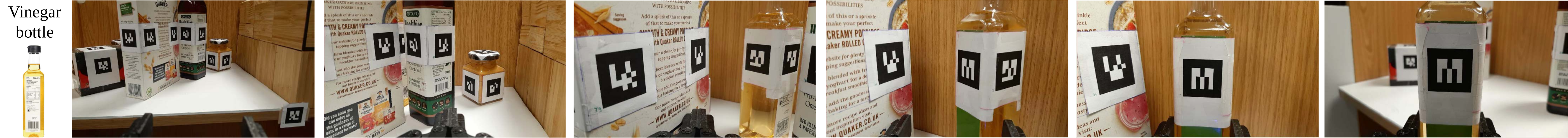} }\vspace{-10pt}
\\
\subfloat[]{ \includegraphics[width=1.0\textwidth]{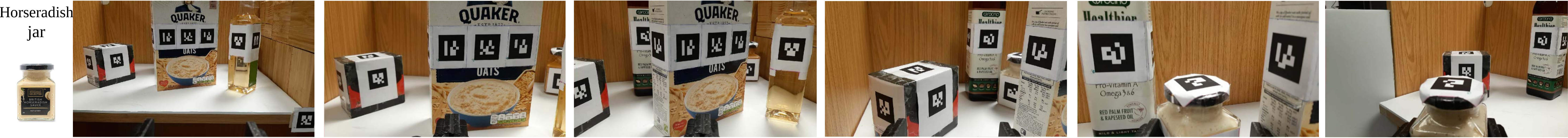} }\vspace{-10pt}
\\
\subfloat[]{ \includegraphics[width=1.0\textwidth]{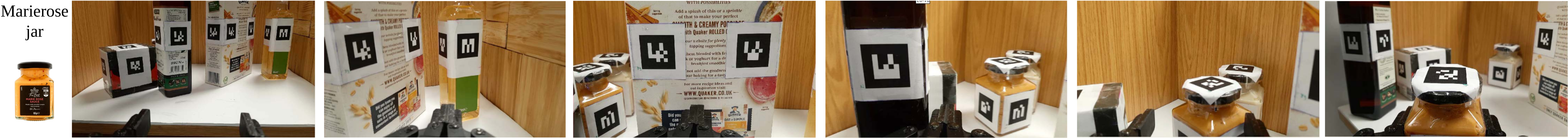}  }
\caption{Snapshots of the hybrid planner retrieving different target objects.}
\label{fig:seq2}
\vspace{-16pt}
\end{figure*}

\textbf{Setup: }
We use two Box2D physics simulators \cite{box2D}, 
one acting as the execution environment 
and the other as the simulation environment where RHP is performed.
The experiments are conducted on
an Intel Xeon E5-26650 computer equipped with an NVIDIA Quadro P6000 GPU.
The experiments evaluate the performance \wrt 
increased clutter density and increased noise level on 
the shape and physics parameters in the execution environment.
The increase in clutter density is aimed at challenging 
the robot with higher occlusion ratios
and more complex multi-object interactions.
The increase in the noise level addresses modelling  errors
between the execution environment and the simulation environment.
Noise is added on the parameters of an object before the execution of an action.
The noise is generated from a Gaussian distribution centred  
around the mean of the object's density $1$ $kg/m^2$ and friction 
coefficient $0.3$. 
Additionally, the shapes of the objects are altered by adding noise on the coordinates of an object's vertices \wrt its centre of mass.
We evaluate the performance over noise levels with standard deviation ranging from $0.0$ to $0.25$ with random number of obstacles up to $10$. 
An experiment with noise level = $0.15$ using Hybrid$_{4\times4}$ is shown in Fig.\ref{fig:noiseSeq}.
The width and depth of the shelf are W:$50$$\times$D:$35\ cm$. 
The dimensions of the gripper are modelled after a \emph{Robotiq 2F-85} gripper
mounted on a \emph{UR5} robot.
\\

\textbf{Results: }
The results are shown in Fig.\ref{exp:sim}.
Each data point in the results is averaged over $300$ task instances with random object configuration.
In terms of success rate, we observe a decreasing trend \wrt clutter density and higher noise levels. 
This is expected as the task becomes more challenging
with higher occlusion ratio and changing dynamics. 
The hybrid planner 
outperforms the other baselines. 
Its success rate improves with higher number of roll-outs and horizon depth as evident by the higher success rate of Hybrid$^{6\times6}$ compared to Hybrid$^{2\times2}$.
Performing a weighted evaluation over the predicted 
poses achieves a slightly higher success rate than just evaluating the most likely one.
Furthermore, the stochastic policies outperform the greedy policy. 
This improvement may be the result of the additional information gained from a stochastic motion.
The stochastic and greedy policies exhibit similar success rates with higher noise levels. 
This is because the changes in physics and object shapes introduce enough randomness in the system
for the greedy policy to act in a similar fashion to the stochastic policy.
The stochastic$_{\mathrm{gen}}$ results are slightly better than its stochastic counterpart, but the difference is not big enough to draw any major conclusion.
The hierarchical planner suffers from the sharpest drop in success rate in both experiments. 
The open-loop execution often fails to produce the intended results.

The average time per task shows a clear advantage for the model-free approaches (greedy, stochastic, and stochastic$_{\mathrm{gen}}$).
Actions are generated almost instantaneously. 
The hybrid planner time degrades with more exhaustive RHP searches.
The difference between Hybrid$^{4\times4}$ and Hybrid$^{4\times4}_{\textrm{lim}}$ 
is not significant despite the latter achieving lower time per task.
This result indicates that the hybrid planner does not often generate a large number 
of potential positions for the target object which would have otherwise resulted
in a bigger time difference.
The hierarchical planner average time is on par with the Hybrid$^{6\times6}$ planner.
These results indicate that simulating the physics during planning 
is the computation bottleneck in a contact-rich environment.

Except for the hierarchical planner, all of the 
approaches perform a similar number of actions per task.
Evidently, the stochastic policies
perform slightly worse than the hybrid planner, 
while the greedy policy is the most efficient.
The hybrid planner, despite relying on stochastic roll-outs, 
executes fewer actions than the stochastic policies as decision 
making is better informed with RHP.
The scale of the number of actions for the hierarchical planer is highly dependent
on the parameters of the underlying low level planners. 
Nevertheless, with a high noise level and clutter density, the high level planner increasingly calls 
the low level planner for re-planning.




%% file: 8.8-Real-World.tex
\subsection{Real-World Experiments} \label{sec:3D}
To validate the simulation results,
we conducted a number of real-world experiments using 
the hybrid planner
for retrieving
a variety of everyday objects (oil bottle, tomato box, jars, oat box, vinegar bottle) from a cluttered shelf.
We used $m$=$4$ and $h$=$4$ as the 
previous section have showed that 
these parameters offer a reasonable balance 
between success rate and execution time.
We mounted an RGB camera
on the end-effector of a \emph{UR5} 
with a \emph{Robotiq 2F-85} gripper 
and manually calibrated the hand-camera transformation. 
We used the Alvar AR tag tracking library for object poses and types detection \cite{Alvar}.
We used Box2D as the simulation
environment to run the hybrid planner.
The shelf dimensions are W:$50\times$D:$35\times$H:$40 \ cm$.
Snapshots from these experiments are shown in 
Fig.~\ref{fig:seq1} and Fig.~\ref{fig:seq2}.
A video of these experiments 
is available on
\textcolor{myblue}{https://youtu.be/dY7YQ3LUVQg}.

We conducted a total of 30 experiments
with up to 8 objects. The robot achieves
a success rate of 90$\%$ with an average of 50 seconds per experiment.
The robot exhibits an informed search behaviour 
by executing actions that increase the
visibility of previously unobserved spaces and by manipulating objects to 
reveal occluded areas behind them. 
In the experiment where the robot is tasked with retrieving the oil bottle,
the robot first approaches the middle of the shelf and searches 
the area behind the oat box. Once the oil bottle is spotted, 
the robot goes around the
cereal box, losing sight of the oil bottle, 
then reaches again for the oil bottle from a less 
cluttered direction.
In the second experiment where the robot is tasked with retrieving the 
vinegar bottle,
we observe the importance of a reactive behaviour when the bottle slips from the
robot's grasp.
The robot is able to recover from this situation by reopening the gripper, approaching
the bottle, and then re-grasping it and pulling it out of the shelf.
In the experiment where the robot is tasked with retrieving the
horseradish jar, the robot pushes obstructing objects to clear an
approach for grasping and retrieving the jar.

Two failed cases were due to the inverse kinematics solver
failing to map an action from the end-effector planar Cartesian space to the robot joint space.
A potential future solution would be to mount 
the robot arm
on a mobile base and leverage the additional gain in
Degrees of Freedom.
Another failure was attributed
to the hybrid planner failing to 
grasp a jar target object
wedged in the corner between the shelf wall 
and 
another obstacle object.


%% file: 9-Conclusion.tex
\section{Conclusions} \label{sec:CONC}
The experiments have shown the efficiency and transferability
of our approach in challenging environments.
The robot's behaviour validates that the
NN stores relevant information from past
observation to guide future actions.
Despite being limited to 2D planar actions,
it offers a stepping stone towards
applications such as object retrieval from fridges 
and supermarket shelves with limited height.

This work forms a solid foundation for extending 
the hybrid planner to 3D manipulations actions 
where the robot can move along the z-axis.
We intend to use tags-free object pose detectors 
in clutter to allow for greater
flexibility~\cite{xiang2017posecnn, wang2019densefusion}.
Additionally,
we envision using an abstract colour-labelled 3D voxelized 
representation of the space with 3D-CNN and 
transformer
architectures.